\documentclass{article}
\usepackage{spconf,amsmath,graphicx}

\usepackage{amssymb}
\usepackage{amsfonts}

\usepackage{graphicx}
\usepackage{subcaption}
\graphicspath{ {images/} }

\usepackage{algorithm,algcompatible}
 
\algnewcommand\INPUT{\item[\textbf{Input:}]} 
\algnewcommand\OUTPUT{\item[\textbf{Output:}]} 

\title{BLIND IMAGE DEBLURRING USING CLASS-ADAPTED IMAGE PRIORS}
\name{Marina Ljubenovi\'{c} and M\'{a}rio A. T. Figueiredo\thanks{The research leading to these results has received funding from the European Union's 
H2020 Framework Programme (H2020-MSCA-ITN-2014) under grant agreement n° 642685 MacSeNet, and was partially supported by the Funda\c{c}\~{a}o para a Ci\^encia e Tecnologia, grant UID/EEA/5008/2013}}
\address{Instituto de Telecomunica\c{c}\~{o}es, Instituto Superior T\'{e}cnico, \\
Universidade de Lisboa, Lisbon, Portugal\\
email: mlju@lx.it.pt and mario.figueiredo@lx.it.pt}
\begin{document}
%
\maketitle
\begin{abstract}
\textit{Blind image deblurring} (BID) is an ill-posed inverse problem, usually addressed by imposing prior knowledge on the (unknown) image and on the blurring filter. Most of the work on BID has focused on natural images, using image priors based on statistical properties of generic natural images. However, in many applications, it is known that the image being recovered belongs to some specific class (e.g., text, face, fingerprints), and exploiting this knowledge allows obtaining more accurate priors. In this work, we propose a method where a Gaussian mixture model (GMM) is used to learn a class-adapted prior, by training on a dataset of clean images of that class. Experiments show the competitiveness of the proposed method in terms of restoration quality when dealing with images containing text, faces, or fingerprints. Additionally, experiments show that the proposed method is able to handle text images at high noise levels, outperforming state-of-the-art methods specifically designed for BID of text images.

\end{abstract}
\begin{keywords}
Blind deblurring, blind deconvolution, ADMM, Gaussian mixtures, plug-and-play.
\end{keywords}
\section{Introduction}
\label{sec:intro}

\textit{Blind image deblurring} (BID) is an inverse problem where the observed image is  modeled as the convolution of an underlying (sharp) image and an unknown blurring filter, often followed by additive noise. The goal of BID is usually to estimate both the underlying image and the blurring filter. The problem is obviously  severely ill-posed. In addition, since the convolution operator itself is typically ill-conditioned, the inverse problem is  highly sensitive to the presence of noise.

In recent years, researchers have investigated a variety approaches to single image BID, mostly considering generic natural images \cite{2006_Fergus_Removing}, \cite{2009_Levin_Understanding}, \cite{2008_Shan_High-quality}, \cite{2009_Cho_Fast}, \cite{2011_Krishnan_Blind}, \cite{2013_Almeida_Blind}. To deal with the ill-posed nature of the BID problem, most methods use prior information on both the image and the blurring filter. The most common choice for the image prior exploits the statistics of natural images \cite{2006_Fergus_Removing}, \cite{2008_Shan_High-quality},  \cite{2009_Cho_Fast}, \cite{2010_Xu_Two-phase}, \cite{2011_Levin_Efficient}, \cite{2011_Krishnan_Blind}, \cite{2013_Xu_Unnatural}  and is usually based on implicit or explicit restoration of salient edges. Although that approach gives good results for natural images, the prior itself is not designed for images that belong to specific classes (e.g., text, face, medical structures, fingerprints) appearing in many important applications, like document analysis, surveillance, and forensics. Methods that use priors that capture the properties of images belonging to specific classes are more likely to provide better results, when dealing with those images, \textit{e.g.}, text \cite{2012_Cho_Text}, \cite{2014_Pan_Text}, \cite{2002_Li_Joint} or face images \cite{2014_Pan_Face}, \cite{2010_Yamaguchi_Facial}, \cite{2011_Huang_Close}. Furthermore, images that belong to different specific classes may have different characteristics that are hard to capture with a unique prior. For example, face images do not contain much texture and text images have specific structure due to the contents of interest being mainly in two tones (commonly, black and white). 

Here, we proposed a method that uses patch-based image priors learned from a set of clean images of the specific class of interest. The method is based on the so-called \textit{plug-and-play} approach, recently proposed in \cite{2013_Venkatakrishnan_Plug}. In contrast with \cite{2013_Venkatakrishnan_Plug}, we do not use a fixed denoiser, but a denoiser based on a Gaussian mixture model (GMM) that is learned from patches of clean images belonging to a specific class. A similar idea was recently proposed for non-blind image deblurring and compressive imaging \cite{2016_Teodoro_Restoration}. Here, in addition to the GMM-based image prior, we also adopt a weak prior on the blurring filter. Considering the blur, earlier methods typically impose hard constrains for the (arguably) most relevant case of a generic motion blur by encouraging sparsity of the blur filter estimate \cite{2009_Cho_Fast}, \cite{2012_Cai_Framelet}, \cite{2006_Fergus_Removing}, \cite{2011_Levin_Efficient}, \cite{2008_Shan_High-quality}, \cite{2010_Wang_Multi-scale}, \cite{2010_Xu_Two-phase}. In this paper, we use a weaker prior on the blur (limited support), thus being able to recover a wide variety of filters then those methods.

\section{Observation model}
\label{sec:observation}

Consider the linear observation model $\textbf{y} = \textbf{H} \textbf{x} + \textbf{n}$, where $\textbf{y} \in \mathbb{R}^{n}$, $\textbf{x} \in \mathbb{R}^{m}$ denote the vectorized (lexicographically ordered) observed data and the (unknown) original image, respectively, and $\textbf{n}$ is noise, assumed to be Gaussian, with zero mean and known variance $\sigma^2$. For computational convenience, $\textbf{H} \in \mathbb{R}^{n \times m}$ is the matrix that represents the convolution with the blurring filter $\textbf{h}$ with periodic boundary conditions, thus with $n = m$. 

As explained above, to deal with the blind image deblurring problem,  prior information (a regularizer)  is imposed on both the underlying image and the blurring filter. The image $\textbf{x}$ and the blurring operator $\textbf{H}$ (equivalently, the filter $\textbf{h}$) are estimated by minimizing the cost function
\begin{equation}
O_\lambda (\textbf{x},\textbf{h}) = \frac{1}{2} ||\textbf{y} - \textbf{H x}||_2^2 + \lambda \phi (\textbf{x}) + \Psi_\textit{S}(\textbf{h}).
\end{equation}
We assume a weak prior on the blurring filter, $\Psi_{\textit{S}}$, the indicator function of the set $\textit{S}$ (set of filters with positive entries on a given support). The rationale behind using a weak prior is that it covers a wider variety of blurring filters.

\begin{equation}
\Psi_{\textit{S}}(\textbf{u}) =   
\begin{cases}
    0       & \quad \text{if } \textbf{u} \in \textit{S} \\
    \infty  & \quad \text{if } \textbf{u} \notin \textit{S}.\\
\end{cases}
\end{equation}
The function $\phi$ represents the prior on the image used to promote characteristics that the underlying sharp image is assumed to have, while parameter $\lambda$ controls the trade-off between data-fidelity term and the regularizer. As shown recently \cite{2010_Almeida_Semi-Blind}, \cite{2013_Almeida_Blind}, good results can be obtained by alternating estimation of the image and the blur kernel (Algorithm 1). Both steps are performed by using the alternating direction method of multipliers (ADMM)  \cite{2013_Almeida_Blind}.

\begin{algorithm}[H]
    \caption{Blind Image Deblurring Algorithm}
  \begin{algorithmic}[1]
    \INPUT Blurred image $\textbf{y}$    
    \OUTPUT Estimated sharp image $\hat{\textbf{x}}$ and the blur kernel $\hat{\textbf{h}}$
    \STATE \textbf{Initialization}: ‎Initial estimate $ \hat{\textbf{x}}  = \textbf{y}$, $\hat{\textbf{h}}$ set to the identity filter, $\lambda > 0$
    \WHILE{stopping criterion is not satisfied}
      \STATE $\hat{\textbf{x}} \gets \underset{\textbf{x}}{\text{argmin}} \hspace{2mm} O_\lambda(\textbf{x}, \hat{\textbf{h}})$
        \COMMENT{estimating $\textbf{x}$ with $\textbf{h}$ fixed}
      \STATE $\hat{\textbf{h}} \gets \underset{\textbf{h}}{\text{argmin}} \hspace{2mm} O_\lambda(\hat{\textbf{x}},\textbf{h})$
      \COMMENT{estimating $\textbf{h}$ with $\textbf{x}$ fixed}
      
    \ENDWHILE
  \end{algorithmic}
\end{algorithm}

In contrast with \cite{2013_Almeida_Blind}, we do not decrease the regularization parameter $\lambda$ at every iteration. We found that a fixed parameter yields better results arguably due to the more expressive prior herein used, when compared with the total-variation regularizer  in \cite{2013_Almeida_Blind}.

\section{ADMM for image inverse problems}
\label{sec:admm}

As discussed above, we use the ADMM optimization algorithm to perform estimation of both the image and the blurring filter, and therefore, in this section, we will briefly explain the ADMM for image inverse problems.  Consider an unconstrained optimization problem in which the objective function is the sum of two functions 

\begin{equation}
\underset{\textbf{z}}{\text{min}} \hspace{2mm}  f_1 ( \textbf{z}) + f_2 ( \textbf{z}).
\end{equation}
By using a variable splitting procedure, we introduce a new variable $\textbf{v}$ as the argument of the function $f_2$, under the constrain that $\textbf{z}=\textbf{v}$. This leads to rewriting the unconstrained problem from above as a constrained one:

\begin{equation}
\underset{\textbf{z}, \textbf{v}}{\text{min}} \hspace{2mm} f_1(\textbf{z}) + f_2 (\textbf{v}) 
\hspace{4mm} \text{subject to} \hspace{4mm} \textbf{z} = \textbf{v}.
\end{equation}
The rationale behind variable splitting methods, such as the method of multipliers or augmented Lagrangian method (ALM), is that it may be easier to solve the constrained problem (4) instead of the unconstrained one (3). 
The main idea behind the ALM is to minimize alternatingly the so-called augmented Lagrangian function

\begin{equation}
\hat{\textbf{z}}, \hat{\textbf{v}} \gets \underset{\textbf{z}, \textbf{v}}{\text{min}} \hspace{2mm} f_1(\textbf{z}) + f_2 (\textbf{v}) 
+ \textbf{d}^T(\textbf{z}-\textbf{v}) + \frac{\mu}{2}||\textbf{z} - \textbf{v}||_2^2,
\end{equation}
and updating the vector of Lagrange multipliers $\textbf{d}$ (Algorithm 2).
In Equation (5), $\mu \geq 0$ is called the penalty parameter. 
If we recall that, by definition, the proximity operator (PO) of some convex function $g$, computed at the point $\textbf{u}$ is defined as

\begin{equation}
\text{prox}_g(\textbf{u}) = \underset{\textbf{x}}{\text{argmin}} \hspace{2mm}\frac{1}{2}||\textbf{x} - \textbf{u}||_2^2 + g(\textbf{x}),
\end{equation}
it is clear that in Algorithm 2, lines 3 and 4 are the PO of $f_1$ and $f_2$, computed at $\textbf{v}^k + \textbf{d}^k$ and $\textbf{z}^{k+1} - \textbf{d}^k$, respectively. 
Formulation (6) can be considered as the solution to a denoising problem, with $\textbf{u}$ as the noisy observation and $g$  the regularizer.


\begin{algorithm}
    \caption{ADMM}
  \begin{algorithmic}[1]
    \STATE \textbf{Initialization}: Set $k = 0$, $\mu > 0$, initialize $\textbf{v}_0$ and $\textbf{d}_0 $ 
    \WHILE{stopping criterion is not satisfied}
      \STATE $ \textbf{z}^{k+1} \gets
      \underset{\textbf{z}}{\text{min}} \hspace{2mm} f_1(\textbf{z}) + \frac{\mu}{2}||\textbf{z} - \textbf{v}^k - \textbf{d}^k||_2^2$    

      \STATE $ \textbf{v}^{k+1}  \gets 
\underset{\textbf{v}}{\text{min}} \hspace{2mm} f_2(\textbf{v}) + \frac{\mu}{2}||\textbf{z}^{k+1} - \textbf{v} - \textbf{d}^k||_2^2$ 

      \STATE $\textbf{d}^{k+1}  \gets   
      \textbf{d}^k - ( \textbf{z}^{k+1} - \textbf{v}^{k+1} )$
     
     \STATE $k \gets k + 1$
    \ENDWHILE
  \end{algorithmic}
\end{algorithm}

\section{GMM-based Denoiser}
\label{sec:gmm}

For the image estimate update (line 3 of Algorithm 1), instead of using the PO of a convex regularizer (line 4 of the Algorithm 2), we implemented a state-of-the-art denoiser considering the fact that the PO itself is a denoising function. This approach, also known as plug-and-play, was recently exploited in \cite{2013_Venkatakrishnan_Plug}, but instead of using a fixed denoiser, such as BM3D \cite{2007_Dabov_Image} or K-SVD \cite{2006_Aharon_KSVD}, we consider a class-adapted GMM-based denoiser \cite{2011_Zoran_Learning}. 

In \cite{2011_Zoran_Learning}, the authors show that clean image patches are well modeled by a GMM estimated from a collection of clean images using the expectation-maximization (EM) algorithm. Furthermore, for a GMM-based prior for the clean patches, the corresponding minimum mean squared error (MMSE) estimate can be obtained in closed form \cite{2015_Teodoro_Single}. We use these facts to obtain a GMM-based prior learned from the set of clean images that belong to the specific class. The rationale behind this approach is that with the class-adapted image prior, we may achieve better performance than with a fixed, generic denoiser, when we process images that do belong to the same specific class. 

\section{Proposed method}
\label{sec:method}

The proposed method uses ADMM for solving each of the inner minimization problems in Algorithm 1 (lines 3 and 4), with $O_\lambda (\textbf{x},\textbf{h})$ as defined in (1) with the PO of $\phi$ replaced by the MMSE estimator using a class-adapted GMM prior. 

\subsection{Image Estimate}
\label{ssec:imageest}

The image estimation problem (line 3 of Algorithm 1) can be formulated as:

\begin{equation}
\hat{\textbf{x}} = \underset{\textbf{x}}{\text{argmin}} \hspace{2mm} \frac{1}{2} ||\textbf{y} - \textbf{H x}||_2^2 + \lambda \phi (\textbf{x}). 
\end{equation}
This problem can be written in the form (3) by setting $f_1(\textbf{x}) = \frac{1}{2} ||\textbf{y} - \textbf{H x}||_2^2$ and $f_2(\textbf{x}) = \lambda \phi (\textbf{x})$.
Applying ADMM to problem (7), yields the so-called SALSA algorithm \cite{2010_Afonso_Fast}. Line 3 of Algorithm 2 becomes a quadratic optimization problem, which has a linear solution:
\begin{equation}
\textbf{x}^{k+1} = (\textbf{H}^T\textbf{H}+\mu\textbf{I})^{-1} (\textbf{H}^T \textbf{y} + \mu (\textbf{v}^k + \textbf{d}^k)).
\end{equation}
As shown in some previous work \cite{2010_Afonso_Fast}, \cite{2011_Afonso_Augmented}, the matrix inversion in (8) can be efficiently computed in the discrete Fourier transform (DFT) domain (using the FFT) in the case of cyclic deblurring, which we consider in this paper. Extension to other boundary conditions can be obtained via the technique proposed in \cite{2013_Almeida_Blind}.

\subsection{Blur Estimate}
\label{ssec:blurest}

The blur estimation problem (line 4 of the Algorithm 1) can be formulated as:

\begin{equation}
\hat{\textbf{h}} = \underset{\textbf{h}}{\text{argmin}} \hspace{2mm} \frac{1}{2} ||\textbf{y} - \textbf{Xh}||_2^2 + \Psi_\textit{S}(\textbf{h}), 
\end{equation}
where $\textbf{h} \in \mathbb{R}^{n}$ is the vector containing the lexicographically ordered blurring filter elements and $\textbf{X} \in \mathbb{R}^{n \times n}$ is the square matrix representing the convolution of the image $\textbf{x}$ and the filter $\textbf{h}$.
Considering formulation (3) we have $f_1(\textbf{h}) = \frac{1}{2} ||\textbf{y} - \textbf{X h}||_2^2$ and $f_2(\textbf{h}) = \Psi_\textit{S}(\textbf{h})$.

The resulting instance of line 3 of  Algorithm 2 has the same form as (8) and, as previously explained, the matrix inversion can be efficiently computed in the DFT domain, using the FFT. Since the proximity operator of the indicator of a convex set is the orthogonal projection on that set \cite{2011_Combettes_Proximal}, line 4 of the Algorithm 2 becomes
\begin{equation}
\text{prox}_{\Psi_{\textit{S}}}(\textbf{u}) = P_{\textit{S}}(\textbf{u}),
\end{equation}
which simply sets to zero all negative elements and any elements outside of the given support.

\section{Experiments}
\label{sec:exp}

In all the experiments, we use the following setting for the two ADMM algorithms described in Section 5: the image estimate is computed with 20 iterations of the algorithm in Subsection 5.1, initialized with the image estimate from the previous iteration, $\textbf{d}_0 = 0$, and $\mu$ hand-tuned for the best visual results or best ISNR (improvement in SNR \cite{2013_Almeida_Blind}) in terms of synthetic data. The blur estimate is computed with two iteration of the algorithm explained in Subsection 5.2, initialized with the blur estimate from the previous iteration, $\textbf{d}_0 = 0$, and $\mu = 0.01$. 

Furthermore, the experiments were performed on three sets of images: (a) a dataset containing 10 \textbf{text} images that is available from the author of \cite{2015_Luo_Adaptive}  (one  for testing and nine for training the mixture), (b) a dataset containing 100 \textbf{face} images from the same author as the text dataset, and (c) a dataset containing 128 \textbf{fingerprints} from the publicly available the UPEK fingerprint database. The GMM-based prior is obtained by using patches of size $6 \times 6$ pixels and a 20-component mixture.

\subsection{Results}
\label{sec:res}

For the experiments with \textbf{text} images, we created five test images using the same clean image of text and $15 \times 15$ synthetic kernels that represent Gaussian, linear motion, out-of-focus, uniform, and nonlinear motion blur, respectively, and noise level corresponding to BSNR = 30 dB (Table 1). For the  \textbf{face} images, we created four $11 \times 11$ synthetic blur kernels that represent Gaussian, linear motion, out-of-focus, and uniform blurs, respectively, and for the fifth experiment, we used blur kernel number 5 from \cite{2009_Levin_Understanding}, with noise level corresponding to BSNR = 40 dB (Table 2). Experiments on the image containg \textbf{fingerprints} are performed with the $15 \times 15$  linear motion blur kernel and noise level corresponding to BSNR = 40 dB (Fig. 2). Results of all experiments are compared with two state-of-the-art BID algorithms constructed for natural images (\cite{2013_Almeida_Blind} and \cite{2011_Krishnan_Blind}), and additionally with the algorithm explained in Subsection 5.1 with the BM3D denoiser plugged into it (PlugBM3D), instead of the GMM-based denoiser (PlugGMM). Note that the generic algorithm \cite{2013_Almeida_Blind} is designed for a wide variety of blur filters, while \cite{2011_Krishnan_Blind}, like the majority of blind deblurring algorithms, is designed mostly for motion blur.

\begin{figure*}[h]

\begin{subfigure}{.2\textwidth}
\centering
\includegraphics[scale=0.2]{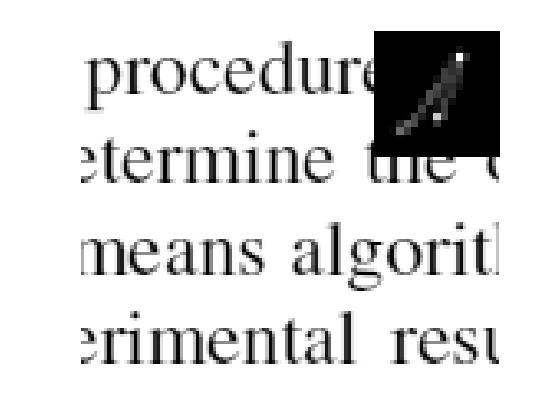}%
\caption{Clean image}
\end{subfigure}%
\begin{subfigure}{.2\textwidth}
\includegraphics[scale=0.2]{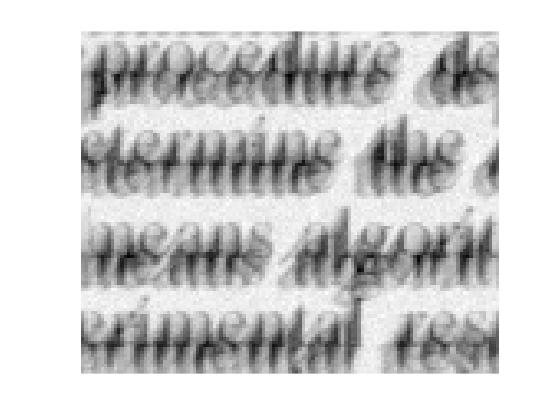}%
\caption{Blurred image}
\end{subfigure}%
\begin{subfigure}{.2\textwidth}
\includegraphics[scale=0.16]{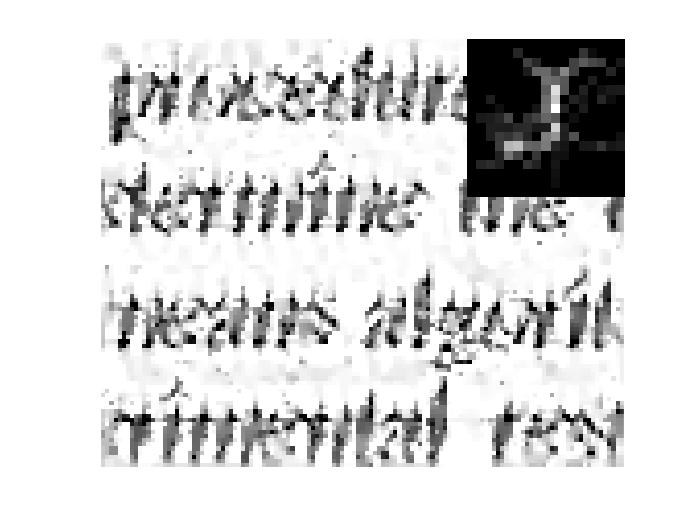}%
\caption{Pan et al. \cite{2014_Pan_Text}}
\end{subfigure}%
\begin{subfigure}{.2\textwidth}
\includegraphics[scale=0.2]{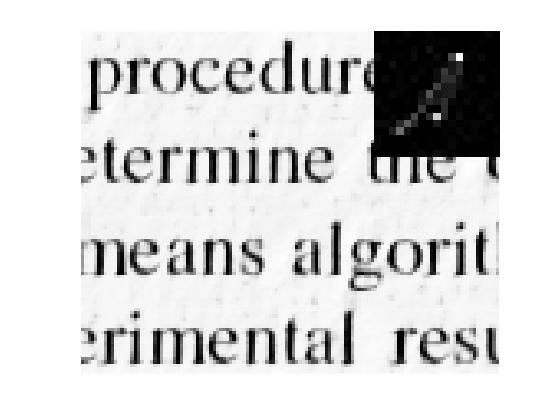}%
\caption{PlugBM3D} 
\end{subfigure}%
\begin{subfigure}{.2\textwidth}
\includegraphics[scale=0.2]{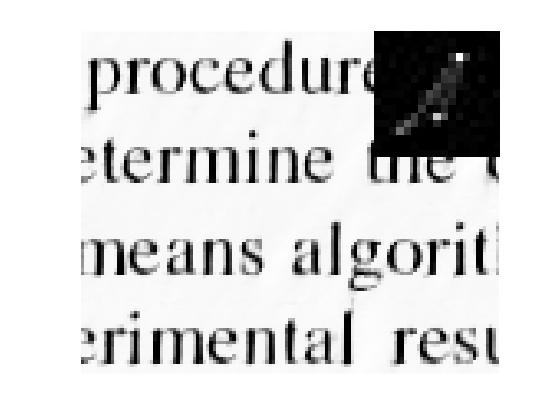}%
\caption{PlugGMM}
\end{subfigure}%

\caption{Text image blurred with nonlinear motion blur number 2 from \cite{2009_Levin_Understanding} and high noise level (BSNR = 20 dB): (a) Original image and ground truth kernel; (b) Blurred image; (c) Results of \cite{2014_Pan_Text}, ISNR = -2.72; (d) PlugBM3D, ISNR = 9.97; (e) PlugGMM, ISNR = \textbf{11.16}.}

\end{figure*}

\begin{figure*}[h]
\begin{subfigure} {.17\textwidth}
\centering
\includegraphics[scale=0.3]{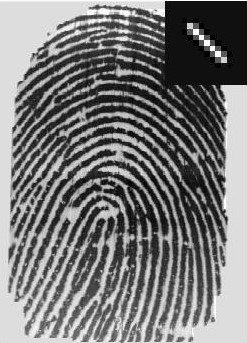}%
\caption{Clean image}
\end{subfigure}%
\begin{subfigure}{.17\textwidth}
\centering
\includegraphics[scale=0.3]{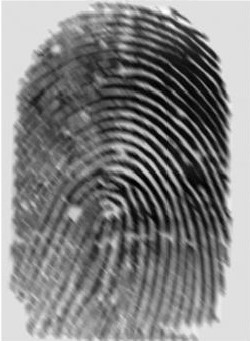}%
\caption{Blurred image}
\end{subfigure}%
\begin{subfigure}{.17\textwidth}
\centering
\includegraphics[scale=0.3]{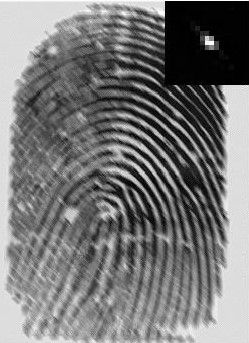}
\caption{Almeida et al. \cite{2013_Almeida_Blind}}
\end{subfigure}%
\begin{subfigure}{.17\textwidth}
\centering
\includegraphics[scale=0.3]{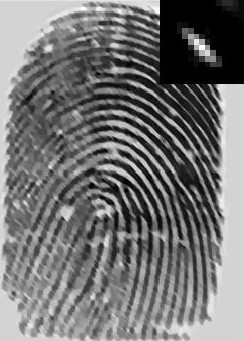}
\caption{Krishnan et al. \cite{2011_Krishnan_Blind}}
\end{subfigure}%
\begin{subfigure}{.17\textwidth}
\centering
\includegraphics[scale=0.3]{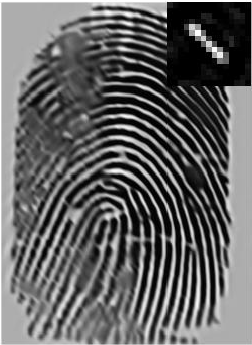}
\caption{PlugBM3D}
\end{subfigure}%
\begin{subfigure}{.17\textwidth}
\centering
\includegraphics[scale=0.3]{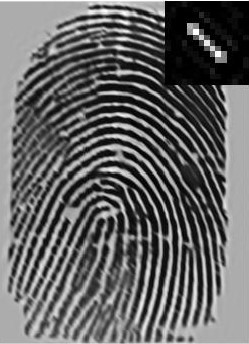}%
\centering
\caption{PlugGMM}
\end{subfigure}%

\caption{Fingerprint image blurred with $9 \times 9$ linear motion blur and noise level (BSNR = 40 dB): (a) Original image and ground truth kernel; (b) Blurred image; (c)  Results of \cite{2013_Almeida_Blind}, ISNR = 0.36; (d) Results of \cite{2011_Krishnan_Blind}, ISNR = -0.64; (e) PlugBM3D, ISNR = 0.56; (f) PlugGMM, ISNR = \textbf{1.19}. }

\end{figure*}

\begin{table}[h]
\centering
\caption{Results in terms of ISNR of the generic methods \cite{2013_Almeida_Blind} and \cite{2011_Krishnan_Blind}, our method using the BM3D denoiser, and our method with the class-adapted GMM prior, tested for \textbf{text} images (BSNR = 30 dB).}
\resizebox{0.48\textwidth}{!}{
\begin{tabular}{cccccc}
\hline
Experiment 	  & 1 & 2 & 3 & 4 & 5\\
\hline
Almeida et al. \cite{2013_Almeida_Blind}      & 0.78 & 0.86 & 0.46 & 0.79 & 0.59\\
Krishnan et al. \cite{2011_Krishnan_Blind}     & 1.62 & 0.12 & - & - & 0.94\\
PlugBM3D     & 7.23 & 8.68 & 8.19 & 8.94 &  13.08\\
PlugGMM      & \textbf{8.88} & \textbf{8.99} & \textbf{9.40} & \textbf{11.48} &  \textbf{16.44}\\
\hline
\end{tabular}}
\end{table}

\begin{table}[h]
\centering
\caption{Results in terms of ISNR of the generic methods \cite{2013_Almeida_Blind} and \cite{2011_Krishnan_Blind}, our method using the BM3D denoiser, and our method with the class-adapted GMM prior, tested for \textbf{face} images (BSNR = 40 dB).}
\resizebox{0.46\textwidth}{!}{
\begin{tabular}{cccccc}
\hline
Experiment 	  & 1 & 2 & 3 & 4 & 5\\
\hline
Almeida et al. \cite{2013_Almeida_Blind}      & 4.31 & 1.81 & 2.86 & 0.85 & 4.43\\
Krishnan et al. \cite{2011_Krishnan_Blind}     & 0.55 & 0.12 & - & - & 0.37\\
PlugBM3D     & 6.64 & 4.86 & 6.78 & \textbf{8.50} &  5.94\\
PlugGMM      & \textbf{7.10} & \textbf{5.30} & \textbf{8.95} & 7.07 &  \textbf{7.33}\\
\hline
\end{tabular}}
\end{table}

Moreover, we tested our method on \textbf{text} images blurred with the blurring filter number 2 from \cite{2009_Levin_Understanding}, followed by a higher noise level (BSNR = 20 dB) (Fig. 1). Results are compared, as previously explained, with the BM3D denoiser plugged into the ADMM loop and the method from Pan et al. \cite{2014_Pan_Text}, which was designed for BID of text images. As the BM3D denoiser is based on exploiting non-local patch similarities, which is highly present in the images we tested, visual results of using PlugBM3D are very good, but in terms of ISNR, PlugGMM clearly outperforms it. 

\section{Conclusion}
\label{sec:conclusion}

In this paper, we have proposed a class-adapted blind image deblurring method, built upon the so-called plug-and-play approach. The method uses Gaussian mixture model (GMM) based denoisers, adapted to specific image classes, plugged into the ADMM optimization algorithm, and a weak prior (positivity and limited support) on the blurring filter. Experiments show that the proposed method yields state-of-the-art results, when applied to images that belong to a specific class (e.g., text, face, and fingerprints), outperforming several generic techniques for blind image deblurring  \cite{2011_Krishnan_Blind}, \cite{2013_Almeida_Blind}. In addition, experiments show that the proposed method can be used for a variety of blurring filters and is able to handle strong noise  in the case of images known to contain text, outperforming the state-of-the-art method for BID of text images \cite{2014_Pan_Text}. 
The proposed method suffers from some potential limitations, such as setting of the regularization parameter and stopping criteria for the inner ADMM algorithms, as well as for the outer iterations, that we aim to improve in  future work.


\bibliographystyle{IEEEbib}
\small
\bibliography{strings,refs}

\end{document}